\documentclass[11pt]{article}
\usepackage[a4paper, margin=1in]{geometry}
\usepackage[utf8]{inputenc}
\usepackage{amsmath,titlesec}
\usepackage{amssymb}
\usepackage{graphicx}
\usepackage{wrapfig}

\title{Homology-constrained vector quantization entropy regularizer}
\author{Ivan Volkov}
\date{ivolk003@ucr.edu}

\begin{document}

\begin{titlepage}
\maketitle
\end{titlepage}

\begin{center} \Large{\textbf{Homology-constrained vector quantization entropy regularizer}}
\end{center}

\begin{center}
\textbf{Abstract}
\end{center}

This paper describes an entropy regularization term for vector quantization (VQ)  based on the analysis of persistent homology of the VQ embeddings. Higher embedding entropy positively correlates with higher codebook utilization, mitigating overfit towards the identity and
\textit{codebook collapse} in VQ-based autoencoders~\cite{jukebox}. We show that homology-constrained regularization is an effective way to increase entropy of the VQ process (approximated to input entropy) while preserving the approximated topology in the quantized latent space, averaged over mini batches. This work further explores some patterns of persistent homology diagrams of latents formed by vector quantization. We implement and test the proposed algorithm as a module integrated into a sample VQ-VAE. Linked code repository provides a functioning implementation of the proposed architecture,  referred to as homology-constrained vector quantization (HC-VQ) further in this work.

\section*{Introduction}

VQ-VAE and derived autoencoders~\cite{https://doi.org/10.48550/arxiv.1711.00937} \cite{https://doi.org/10.48550/arxiv.1906.00446} can efficiently compress high-dimensional data while preserving the discrete nature of compressed features.

Vector quantization VAEs are prone to codebook collapse, a phenomenon of vanishing codebook utilization and vanishing entropy, which has negative effects on robustness of dimensionality reduction and number of features learned~\cite{wav2vec}. In some works it is addressed by random restarts of\textit{ dead }(inactive) codes ~\cite{jukebox}. In this paper we propose a controlled entropy regularization mechanism through topology  adjustments of the quantized latent space.

As described previously, periods of decreased VQ entropy appear commonly throughout training. The phenomenon is similar in nature to posterior collapse observed in VAEs. This constitutes the unwanted convergence of latent space $q_\phi(z \mid x)$ towards the prior such that a VAE effectively stops receiving any new informative features from the input data: $\exists i$ s.t. $\forall \mathbf{x} q_\phi\left(z_i \mid \mathbf{x}\right) \approx p\left(z_i\right)$

Heuristically, overly large latent entropy with a high $\sigma$ over an interval may be a good indicator that the model is not converging, and struggling to find effective gradients for parts of the computation graph. Hence it makes sense to explore approaches of deriving entropy-inducing mechanism that would be structural, upper-bounded and would not undermine overall learning stability.

The research was motivated by the theory of entropy coding, which was developed in the context of data compression. A classical work in signal processing that made the relation between entropy coding and vector quantization -  ECVQ~\cite{ecvq} - proposed entropy-constrained vector quantization for the analog signal. We found the core idea behind this analog signal algorithm can be translated to dimensionality reduction performed in modern autoencoder architecture.
In that regard, parts of data compression theory (specifically, entropy coding and source coding theorem) are applied to VQ while dimensionality reduction itself is interpreted as a \textit{stochastic} data compression process in this context.

More formally, entropy coding is a process of compressing data while minimizing code (representational unit) length within limit of the lower bound of Shannon source coding theorem~\cite{source-coding} that defines bounds of lossless compression \textit{wrt} the entropy of the source data:
\\
$$\mathbb{E}_{x \sim P}[l(d(x))] \geq \mathbb{E}_{x \sim P}\left[-\log _b(P(x))\right]$$
\\
where $d$ is the coding function, $l$ is the number of symbols in a code word, $b$ is the number of symbols used in output codes, and $P$ is the probability of the source symbol.

We make an assumption that we can generalize the theorem to the VQ codebook $\mathcal{X}$  utilization rate as opposed to individual code length (as each element of the codebook has a fixed representational capacity), given that $l(d(x)) \in \mathcal{X}$. In other words, when we project the source coding theorem from entropy coding to VQ, our assumptions are: (1) higher codebook entropy corresponds to a more feature-rich  (\textit{lossless}) dimensionality reduction but can lead to decreased learning speed, while  (2) lower codebook entropy corresponds to a faster learning rate (faster reconstruction convergence in VQ-VAE) but can lead to a \textit{lossier} dimensionlaity reduction.

However, we found that simply targeting increased codebook entropy undermines learning stability and meaningful gradient computation. While experimenting with various constrained entropy regularization methods we came to observe some persistent patterns linking entropy to particular topological features found in derived persistent homology. We found that a differentiable implemention of PH can be used to define topological parameters which invoke consistent entropy feedback without undermining the general stability of learning \textit{wrt} to autoencoder reconstruction loss.

Subsequently, in this work we explore an entropy regularization term for VQ process in the context of VQ-VAE, based on the analysis of generated persistent homology features.

\section{Related work}

One of the major relevant works is the \textit{topological autoencoder}~\cite{topo-ae}, where a loss term \textit{wrt} $ \left(\mathbf{A}^X, \mathbf{A}^Z, \pi^X, \pi^Z\right) $ was proposed, corresponding to the distance matrices $A$ and topologically relevant persistent pairings $\pi$ of input space $X$ and latent space $Z$, respectively. Proposed topological autoencoder has proven to find consistent patterns in observed persistent homology (PH) of constructed Vietoris-Rips filtrations, strengthening the assumption that persistent homology can be applied to dimensionality reduction.

The research was also motivated by a work that researched PH learning ~\cite{can-hom-be-learned}, investigating whether neural networks are able learn PH features of $F \circ \mathrm{PH}: \mathcal{X} \rightarrow \mathcal{D} \rightarrow \mathcal{Y}$ directly, where $\mathcal{X}$ is the space of inputs, $\mathcal{D}$ is the space of persistence diagrams, and $\mathcal{Y}$ is the space of PH features. The results indicated that a CNN can efficiently approximate various types of persistence diagrams.

 Two works addressing VQ codebook collapse - referred to as wav2vec~\cite{wav2vec} and Jukebox~\cite{jukebox} after their model labels - pushed the research to further link codebook collapse and VQ entropy. We also make an assumption that higher entropy of VQ latents correlates inversely with (unwanted) overfit towards the identity \cite{ae-identity-overfit}, systemically observed in autoencoders, usually addressed by entropy-inducing measures such as dropout and denoising mechanism.

 Ideas of correlating entropy coding to analog signal quantization, as mentioned, are presented in the ECVQ work that was published as early as 1989~\cite{ecvq}. Success of the ECVQ design - where minimum distortion was effectively subjected to entropy constraints - sets the theoretical basis of our entropy-constrained VQ design.

 Finally, we borrow an analytical term (with the corresponding formula) from a work that defined persistent entropy~\cite{persistent-entropy}. Persistent entropy refers to the Shannon entropy of the filtration $\mathcal{D}$, in other words it's an entropy measure of the persistent homology itself. Persistent entropy is interpreted as the degree of topological order in persistent homology, useful for discerning between a noisy diagram and a diagram containing structural topological features.

\section{HC-VQ Entropy Regularization}

\textbf{Persistent homology of latent space} Johnson-Lindenstrauss lemma (JL-lemma) is central to theory backing dimensionality reduction. JL-lemma defines  a practical distortion limit of  $(1 \pm \varepsilon)$ for any Euclidean point cloud of \textit{n} points when embedded into a space with $O\left(\varepsilon^{-2} \log n\right)$ dimensions.

It has been shown that persistent homology of a point cloud can also be approximately preserved with similar constraints: random projections approximately preserve persistent homology up to $1 \pm \epsilon$ for $\varepsilon \in[0,1]$~\cite{homology-preserved-0}~\cite{homology-preserved-1}, assumption tested in practice in the topological autoencoder experiment~\cite{topo-ae}.

Assumption of preserved persistent homology becomes foundational for this work, as we can induce that a compressed representation (less generally, quantized latents) can be equipped with an \textit{``optimal''} persisent homology which would be representative of  features of the  higher-dimensional data.

\textbf{Constructing persisent homology for embeddings} We examined vector quantization as implemented in a VQ-VAE, thus many aspects are best understood in the context of autoencoders and deep learning. However, the algorithm can be generalized to vector quantization independent of deep learning process.

On a higher level, it makes sense to apply topological analysis to deeper autoencoder layers, as dim. reduction moves representations from input-specific perturbations of the ``shallow layers'' (e.g. image noise) to more general feature layers of the bottleneck.

VQ layers are situated at or close to autoencoder bottleneck, working on data preprocessed by shallow layers and use k-NN algorithm to quantize their latents in the following fashion:
\begin{equation}
z_q(x)=e_k, \quad \text { where } \quad k=\operatorname{argmin}_j\left\|z_e(x)-e_j\right\|_2
\end{equation}

As per classical VQ-VAE implementation, an embedding table weight serves as a map of 1-nearest neighbors, to which tensors of input are quantized. We perform our topological analysis on the embedding table domain \textit{wrt} the entropy of the embeddings, constructing a Vietoris-Rips filtration with respective persistent homology as follows.

Let $(\mathcal{X}, d)$ be the assigned metric space of the $\mathcal{X}$ embedding table point cloud. The associated Vietoris-Rips complex of $\mathcal{X}$ is defined as $V R_s(\mathcal{X})$ with a distance threshold of $\epsilon$:
$$
V R_s(\mathcal{X})=\left\{\left[v_0, \ldots, v_n\right] \mid \forall i, j d\left(v_i, v_j\right) \leq \epsilon\right\} .
$$

The persistent simplicial cohomology of the constructed VR complex $H^{*}(\mathcal{X}(\epsilon))$ is obtained by assembling a disjoint set of simplicial complexes (filtration) at various $\epsilon$ in (0, $k$) dimensions - for the purposes of this work we define $k$ at 1:
$$K_{\epsilon_0} \longrightarrow K_{\epsilon_1} \longrightarrow K_{\epsilon_{n-2}} \rightleftarrows K_{\epsilon_{n-1}}$$
$$\begin{aligned}&H^{k}\left(K_{\epsilon_0}\right) \longrightarrow H^{k}\left(K_{\epsilon_1}\right) \longrightarrow \cdots \longrightarrow H_k\left(K_{\epsilon_{n-2}}\right) \longrightarrow H_k\left(K_{\epsilon_{n-1}}\right) \\&\end{aligned}$$\

\textbf{Homology signature} Having obtained $H^{k}(\mathcal{X}(\epsilon))$ and it's corresponding persistent homology diagram $\mathcal{D}$, we can investigate correlations of some aspects of $\mathcal{D}$ \textit{wrt} selected entropy parameters emerging during VQ-VAE training. For our choice of $k$, we found two particularly sensitive features of $\mathcal{D}$, which will define the partial persistent diagram signature $\mathcal{ S^{\mathcal{D}}}$ for the purposes of this work:
$$\mathcal{ S^{\mathcal{D}}} := \{\beta^{1}, \mu^{0}\}$$

Where $\beta^{1}$ refers to the 1-dimensional Betti numbers (which can be intuitively viewed as the number of loops formed by enclosing 1-simplices of the filtration), and where $\mu^{0}$ refers to the mean distance between critical 0-simplices (which can be intuitively viewed as a sparsity metric between connected components of the filtration).

\textbf{Entropy correlation} We measure entropy of quantized latents with the standard entropy formula~\cite{jukebox}. We evaluate the entropy of quantized latents (for each mini batch), not just the entropy of the codebook - such that the model can learn data-specific regularization of the embeddings.
$$E_{codebook} = -\sum_{i=1}^n (K_{p} * \log _e\left(K_{p}\right))_{i},\ \ \   where\ K_{p} = \dfrac{z_e(x)}{\sum_{i=1}^n (z_e(x))_{i}} $$
Where $z_e(x)$ denotes a vector space of quantized latents. We further evaluate it against $E(H)$ where $H$ is the input data.

In addition, we analyze the persistent entropy of the filtration, however we do not include it in our loss calculation. Persistent entropy $E(F)$ definition is taken from the previously cited work~\cite{persistent-entropy}:

$$E(F)=-\sum_{i=1}^n p_i \log \left(p_i\right) \text { where } p_i=\frac{\ell_i}{S_L}, \ell_i=y_i-x_i \text {, and } S_L=\sum_{i=1}^n \ell_i \text {. }$$
Let $F$ be a (Vietoris-Rips simplicial) filtration constructed from the point cloud $\mathcal{X}$ and the corresponding persistence diagram $\operatorname{\mathcal{D}}(F)=\left\{a_i=\left(x_i, y_i\right) \mid 1 \leq i \leq n\right\}$ (being $x_i<y_i$ for all i), let $L=\left\{\ell_i=y_i-x_i \mid 1 \leq i \leq n\right\}$.

\textbf{Regularization term} We define the dynamical HC-VQ regularization loss term \textit{wrt} to the $E_{z}$ entropy, where $z$ refers to the point cloud of quantized latents. The term is scaled by an arbitrary domain-specific hyperparameter $\lambda$ for numerical stability:
\\
$$\mathcal{L}_{overall}:= \mathcal{L} + \mathcal{L}_r(\mathcal{ S^{\mathcal{D}}}, \lambda*\Delta E_{z}/ E_{input}*T_{\mathcal{S}})$$

$$\ where\ \Delta E_{z}\ := \ \mu_{k}\{E_{z}\}$$
\\

We accumulate entropy count $E_{z}$ for each step (mini batch) and then calculate a mean value as a measure of entropy over a step interval $k$. More specifically,  $k$ is a\textit{ period }hyperparameter referring to a number of training steps - or, more generally, vector quantization attempts to average on. Higher $k$ value produces smoother entropy estimations.

Target value of $ S^{\mathcal{D}}$ is defined by two hyperparameters $T_{\beta}, T_{\mu}$ comprising the target homology signature $T_{\mathcal{S}}$.

\section{Implementation and Hyperparameters}

A functioning implementation of HC-VQ module integrated to a VQ-VAE is made available in a public git repo.~\cite{git-hcvq} We use an embedding table $X \in \mathbb{R}^{128 \times 64}$. An embedding layer is incorporated at the bottleneck dimension of a VQ-VAE with 2 convolutional and 2 ReLU layers at each side of the autoencoder.

After quantizing the latents, we calculate step entropy $E_{z}$ and compare it to the entropy of the source $E_{input}$ at each $k = 200$.

HC-VQ is implemented with full support for backpropagation: construction of persistence diagrams is differentiable; Betti numbers are counted using approximate floating-point counting based on narrow gaussians.

Loss is calculated with the standard MSE function, and returned as part of the overall VQ module loss. Learning rate is set to $0.0008$, VQ-VAE commitment loss $\beta$ is set to 0.25.

The base VQ-VAE implementation is reused from a sample VAE autoencoder collection repository.\cite{vae-git}
Vietoris-Rips filtration is calculated with giotto-ph software that supports calculation of gradients. We count Betti numbers without thresholding of $\epsilon$, such that all the returned diagram pairwise distance are counted.
 $k$-value - number of training steps for the $E_{z}$ aggregation period - is set to 200.

\section{Experiments}

We test our HCVQ in the context of a sample VQ-VAE on three datasets with varying complexity: CelebA face dataset, MNIST number dataset, and Fashion-MNIST articles of clothing dataset.

We train two VQ-VAE models for 27000 steps each, one with HC-VQ term deactivated and one with HC-VQ term activated (participating in overall loss computation). Same dataseed is reused for producing predictable results of stochastic or pseudorandom functions.

\textbf{ HC-VQ dry runs } The following observations were made during the most complex (CelebA) dataset training. In $\mathcal{S^{\mathcal{D}}}$, $\mu^{0}$ (mean $\epsilon$ between persistent pairings of critical simplices in dimension 0) was most correlated to entropy $\Delta$s at each interval. $\mu^{0}$ was increasing steadily as training went on:

\begin{figure}[htp]
    \centering
    \includegraphics[width=10cm]{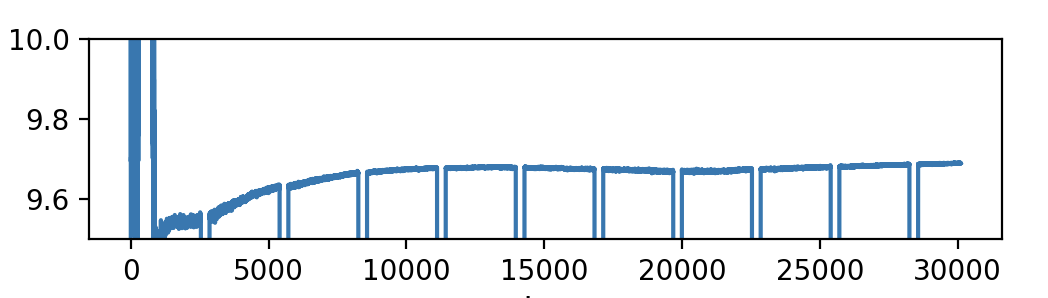}
    \caption{$E_{z}$ entropy with no HCVQ}
    \label{fi:s}
\end{figure}

\begin{figure}[htp]
    \centering
    \includegraphics[width=10cm]{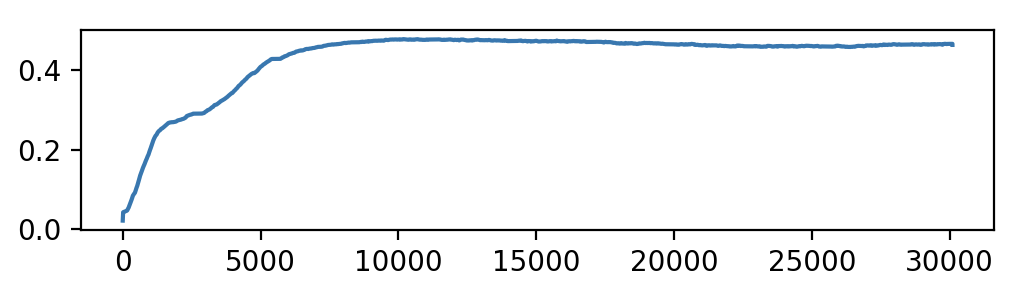}
    \caption{$\mu^{0}$ with no HCVQ}
    \label{fi:z}
\end{figure}

Figure 1 presents an instance of diminished codebook utlization ($\approx(0, 5000)$ step interval)
\\


\textbf{ Comparisons to active HC-VQ term} The following were CelebA training observations with HCVQ loss term active. A set of hyperparameters \{$T_{\beta} := 100 $, $T_{\mu} := 0.5$\} influenced the $E_{z}$ entropy parameter rather drastically:
\\
\begin{figure}[h!]
    \centering
    \includegraphics[width=10cm]{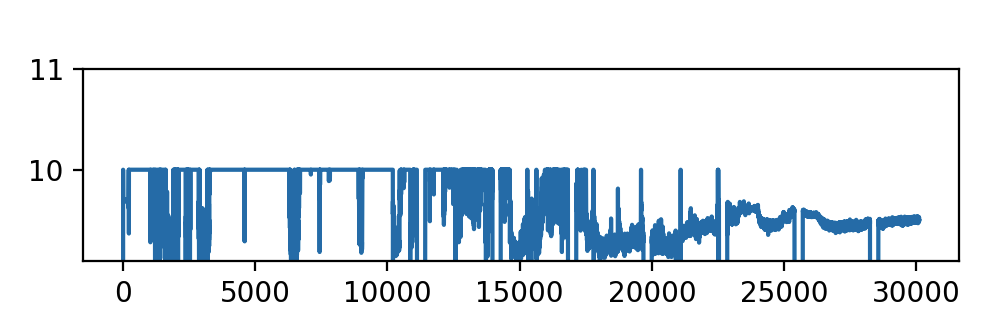}
    \caption{$E_{z}$ entropy with HC-VQ, unstable $E > 10$ values are clamped}
    \label{fi:zz}
\end{figure}
\\
\\
\\
By tracking the relaionship of reconstruction loss and entropy we can observe some periods of model instability where it was struggling to fit the nudges of the HC-VQ:
\\
\\
\\
\\
\\
\\
\\
\\
\\
\\
\\
\\

\begin{figure}[h!]
    \centering
    \includegraphics[width=10cm]{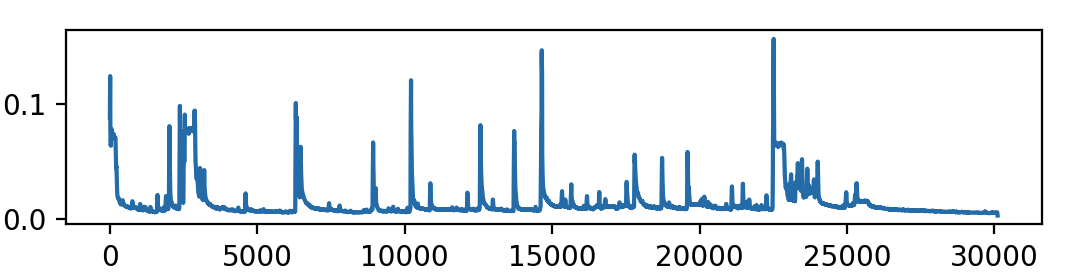}
    \caption{Reconstruction loss with HC-VQ}
    \label{fi:ae}
\end{figure}

And a respective $\Delta E_{z}/ E_{input}$  entropy loss metric:

\begin{figure}[h!]
    \centering
    \includegraphics[width=10cm]{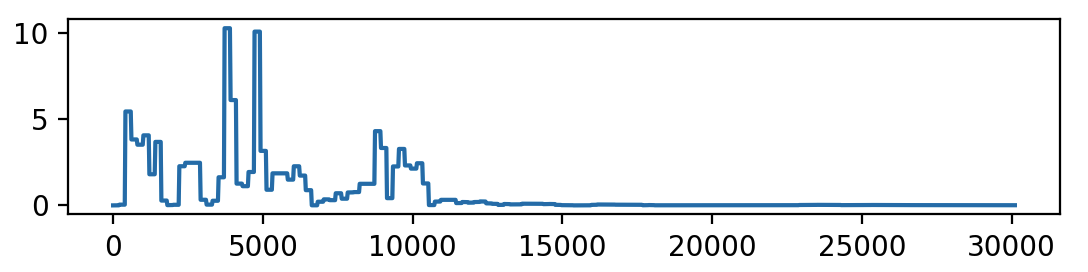}
    \caption{HC-VQ $\Delta E_{z}/ E_{input}$ $\mathcal{L}$}
    \label{fioo:s}
\end{figure}

Persistent entropy indicates a more orderly topological structure with HC-VQ enabled:

\begin{figure}[h!]
    \centering
    \includegraphics[width=8cm]{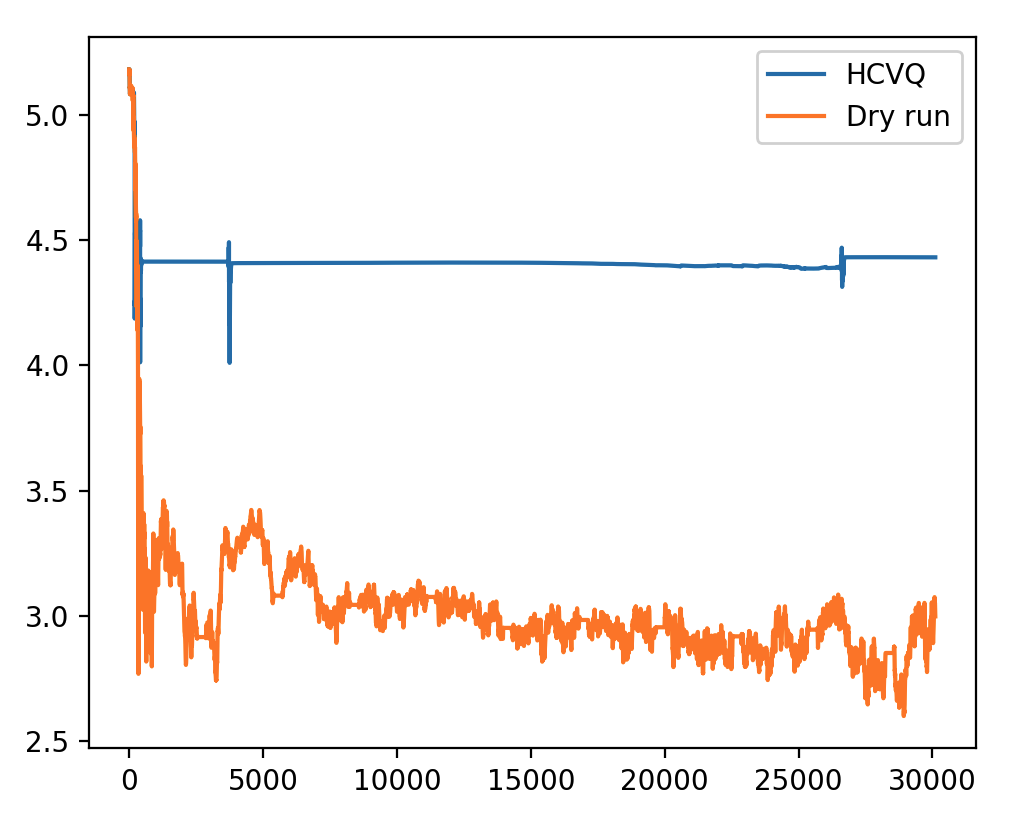}
    \caption{Persistent entropy comparison}
    \label{fixx:s}
\end{figure}

A snapshot of final diagrams (which remained stable over last $k$-intervals) indicates a more sparse positioning of connected simplices with HC-VQ enabled:

\begin{figure}[h!]
    \centering
    \includegraphics[width=10cm]{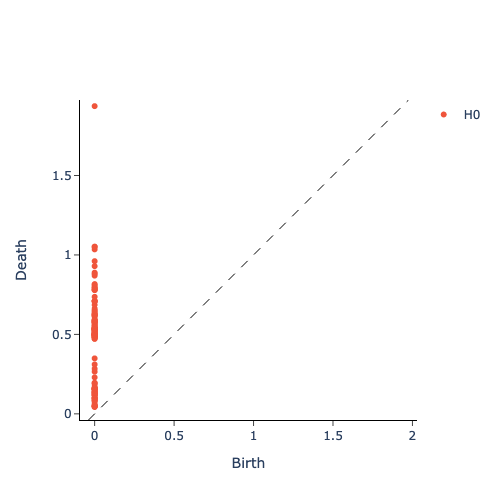}
    \caption{Dry-run: step 27000 persistent homology diagram}
    \label{ficcxx:saa}
\end{figure}
\textbf{}
\begin{figure}[h!]
    \centering
    \includegraphics[width=10cm]{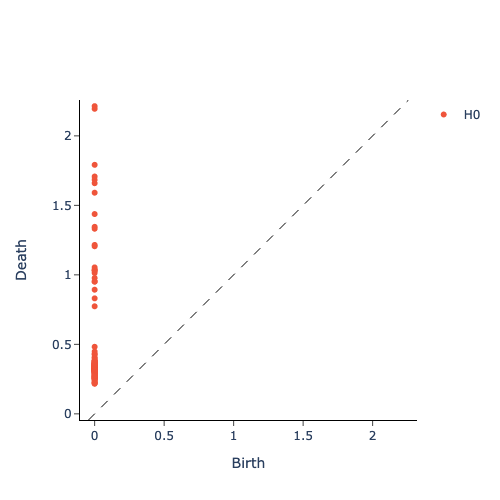}
    \caption{HC-VQ active: step 27000 persistent homology diagram}
    \label{fixhh:sdd}
\end{figure}

\section{Conclusion}

We found that HC-VQ term technique was able to efficiently increase codebook utilisation (measured by $E_{z}$) during VQ-VAE training process compared to dry runs. Heuristically, we observed sparse persistent homology and (stable) high persistent entropy, strengthening the assumption that HC-VQ embedding retained more features compared to VQ-VAE training with the same codebook size.

The biggest drawback of HC-VQ would be numerical stability, which can be mitigated by fitting an appropriate $\lambda$ scaling hyperparameter and decreasing hyperparameters $T_{\beta}, T_{\mu}$ (at the cost of decreased learning speed).

\bibliographystyle{IEEEtran}
\bibliography{bibl}
\end{document}